\documentclass[conference]{IEEEtran}
\IEEEoverridecommandlockouts
\usepackage{cite}
\usepackage{amsmath,amssymb,amsfonts}
\usepackage{algorithmic}
\usepackage{graphicx}
\usepackage{textcomp}
\usepackage{xcolor}
\usepackage{natbib}
\usepackage{hyperref}
\usepackage{caption}
\usepackage{multirow}
\usepackage{subcaption}

\def\BibTeX{{\rm B\kern-.05em{\sc i\kern-.025em b}\kern-.08em
    T\kern-.1667em\lower.7ex\hbox{E}\kern-.125emX}}

\begin{document}

\title{Revisiting the Role of Label Smoothing in Enhanced Text Sentiment Classification}


\author{\IEEEauthorblockN{Yijie Gao \qquad Shijing Si$^{*}$\thanks{$^{*}$Corresponding Author: Shijing Si, shijing.si@outlook.com} \qquad Hua Luo \qquad Haixia Sun \qquad Yugui Zhang}
\IEEEauthorblockA{\textit{School of Economic and Finance} \\
\textit{Shanghai International Studies University}\\
Shanghai, China }}

\maketitle

\begin{abstract}
Label smoothing is a widely used technique in various domains, such as text classification, image classification and speech recognition, known for effectively combating model overfitting. However, there is little fine-grained analysis on how label smoothing enhances text sentiment classification. To fill in the gap, this article performs a set of in-depth analyses on  eight datasets for text sentiment classification and three deep learning architectures: TextCNN, BERT, and RoBERTa, under two learning schemes: training from scratch and fine-tuning.
By tuning the smoothing parameters, we can achieve improved performance on almost all datasets for each model architecture. We further investigate the benefits of label smoothing, finding that label smoothing can accelerate the convergence of deep models and make samples of different labels easily distinguishable. 
\end{abstract}


\section{Introduction}

Text sentiment classification is to identify and extract emotional tendencies (such as positive, negative or neutral) from text, which provides important application scenarios and challenges for the development of natural language processing (NLP) \citep{li2022survey}. Many methods have been
proposed for this task, for instance,
convolutional neural networks (CNN) \citep{zhang2015character}, and recurrent neural networks (RNN) with attention mechanism  \citep{zhang2016sentiment,chen2017recurrent}. Transformer-based models  \citep{jiang2021lightxml,chang2020taming} have excelled in sentiment classification tasks, achieving state-of-the-art performance  \citep{liu2022co,liu2022combining}. 

As a commonly used regularization method to overcome overfitting,
label smoothing (LS) proceeds by using soft targets that are a weighted average of the hard targets and the uniform distribution over labels  \citep{muller2019does,lienen2021label}.
This technique lessens the disparity between the top probability estimate and the remaining ones, thereby acting as a barrier to the model from generating extremely confident predictions, consequently decreasing the model's likelihood of becoming excessively tailored to the training data  \citep{lukasik2020does,gao2020towards}. This also fosters generalization, leading to better performance on unseen data  \citep{chen2020investigation,cui2022improving}. Furthermore, LS can also mitigate the impact of noisy labels on the training process \citep{pmlr-v108-li20e}. 
 
LS has achieved widespread success in NLP  \citep{chandrasegaran2022revisiting,liu2022label} by introducing soft targets, which allows the model to optimize towards a more flexible direction during training \citep{haque2023label,pan2022efficient}.
However, the fine-grained research on LS for text
sentiment classification is still limited.
For the task of sentiment classification, emotion is not an absolute and discrete concept  \citep{pang2002thumbs,ONAN20222098}, but there is certain fuzziness and continuity \citep{hasib2021online}. LS can take account of this ambiguity, allowing the model to learn more balanced relative probabilities between different emotions.

 In order to investigate how LS benefits sentiment classification, in this paper we conduct extensive experiments on three widely used deep neural network architectures: TextCNN, BERT, and RoBERTa, under two learning schemes: training from scratch and fine-tuning. Our contributions can be summarized as follows:
 \begin{itemize}
     \item  Through varying the tuning smoothing parameter, LS methods outperform the three baseline architectures on all eight datasets, including six three-class datasets and two binary-class datasets.
     \item From in-depth analysis, LS can accelerate the training process of deep models with the deployment of soft labels
     \item LS can produce better hidden representations for training examples as they are easier to distinguishable than those produced by the baseline method.
 \end{itemize}

The remaining sections of this paper are organized as follows. We first review the related works in Section 2. Then, in Section 3, we propose the application and deployment of LS in text sentiment classification. Subsequently, in Section 4, we present the experimental results on eight sentiment analysis datasets. Finally, Section 5 provides a discussion and conclusion of this paper.

\section{Related Works}

This paper is related to two lines of research: text 
sentiment classification and label smoothing.

\subsection{Text Sentiment Classification}

Text sentiment classification is one special type of text classification  \cite{kowsari2019text,si2020students}, 
as the labels are ordinal.
In early days, statistical-based models like Naive Bayes  \citep{kim2006some,raschka2014naive} and support vector machines  \citep{tong2001support,lilleberg2015support} were dominant in text sentiment classification. These models, while accurate and stable, required time-consuming feature design and often overlooked the context information in text data \citep{kowsari2019text}.
Since the 2010s, deep learning models  \citep{zulqarnain2020comparative,liu2016recurrent}, which automatically extract meaningful representations for text without the need for manual rule and feature design \citep{minaee2021deep}, are increasingly utilized for sentiment classification. 

Although deep learning methods have achieved good performance on text sentiment classification, they may still suffer issues like slow convergence, poor performance.
In this paper, we explore how LS benefits deep sentiment methods.

\subsection{Label Smoothing (LS)}
LS is firstly introduced in the field of computer vision (CV) and has achieved success in various visual recognition tasks \citep{Gao2017,muller2019does, si2022towards}. Later, the method is shown to be effective in Machine Translation (MT) \citep{gao2020towards}. Moreover, it also has applications in sentiment classification \citep{luo2022early,luo2021smoothing} and Named Entity Recognition(NER) \citep{zhu2022boundary,yu2022chinese}, improving model calibration and bringing flatter neural minima.

Meanwhile, researchers have started to apply LS to text sentiment classification. One attempt is to add LS to the loss function to enhance the performance of the model \citep{wang2022sentiment}. Similarly, another group add LS to loss function and perform emotion classification on the adaptive fusion features obtained \citep{yan2023polarity}. Also, properties of LS and its adversarial variants are studied, showing LS can enhance the adversarial robustness of the model \citep{yang2022and}.  \citet{yan2023polarity} proposes a new cyclic smoothing labeling technique for handling the periodicity of angles and increasing error tolerance for adjacent angles. They also designed a densely-coded tag that greatly reduced the length of the code.  \citet{wu2022text} propose an efficient data augmentation method, termed text smoothing, by converting a sentence from its one-hot representation to a controllable smoothed representation and show that text smoothing outperforms various mainstream data augmentation methods by a substantial margin.

Although LS is a common label processing technique, there are still some problems and challenges that need further study. Examples include comparison of effects across different data sets, tasks, and model structures, comparison with other regularization methods, etc \citep{huang2020learning}. Studying LS can explore these problems and promote the further development of the field of LS. Our work revisits the application of LS to sentiment classification, and conducts an in-depth analysis of its power. 

\section{Label Smoothing Method for Text Classification}\label{sec:method}

Given a tokenized input $\boldsymbol{x} = (x_{1}, x_{2}, \ldots, x_{n})$ and a set of labels $\boldsymbol{y} = (y_{1}, y_{2}, \ldots, y_{k})$, where $n$ is the length of input and $k$ is the number of categories for classification. Thus, $\boldsymbol{D}_{i} = (d_{x_{i}}^{y_{1}}, d_{x_{i}}^{y_{2}}, \ldots, d_{x_{i}}^{y_{k}})$ describes the extent to which the instance $x_{i}$ belongs to a label $y_{k}$. We expect to obtain a final classification result $d_{i}$ which is typically the maximum value of the label distribution $\boldsymbol{D}_{i}$ obtained by applying a softmax normalization to the last layer of the deep neural networks. The label corresponding to the maximum value represents the category to which this document belongs.

\subsection{Basics of Label Smoothing}



Compared with the commonly used hard target, label distribution has some advantages.
Due to the continuity of label distribution, it has a larger labeling space, a broader range of expression, and therefore can provide greater flexibility in the learning process. Specifically, LS is a straightforward way to convert hard labels to soft label distributions, which is a mixture of the one-hot hard label and the uniform distribution, i.e., $\boldsymbol{D}_{i}^{'} = (1-k\lambda)\boldsymbol{D}_{i} + \lambda\boldsymbol{1}, \lambda \in [0, 1]$, where the original one-hot distribution $\boldsymbol{D}_{i} = (d_{x_{i}}^{y_{1}}, d_{x_{i}}^{y_{2}}, \ldots, d_{x_{i}}^{y_{k}})$, $\boldsymbol{1}=(1, 1, \ldots, 1)_{1\times{k}}$, and $\lambda$ is the smoothing parameter.

This method also addresses the need for manual labeling. One of the primary challenges in the task of LS is the difficulty of obtaining the true label distribution. Most classification datasets do not provide this information, and theoretically, acquiring the precise label distribution would require extensive manual labeling of the same sample to obtain its statistical distribution, which is prohibitively expensive.

Suppose that $\theta$ is the trainable parameters of a classification model, and for the $i$-th training example, $P(\boldsymbol{y}|x_{i}; \theta)=p(y_{1}|x_{i}; \theta), p(y_{2}|x_{i}; \theta), \ldots, p(y_{k}|x_{i}; \theta)$ is the probability distribution finally output by the model. We use Kullback-Leibler divergence during the training process,that is, the optimal parameter $\theta^{*}$ satisfies:

\begin{equation}
\begin{split}
\theta^{*} &= \arg \min\limits_{\theta} \sum_{i}KL(\boldsymbol{D}^{'}_{i}||P_{i})\\
&= \arg\min_\theta \sum_{i} \sum_{j} \left( d^{'y_j}_ {x_i} \ln \frac{d^{'y_j}_{x_i}}{p(y_{j}|x_{i}; \theta)} \right)
\end{split}
\end{equation}

This expression can be transformed into a maximum likelihood function:
\begin{equation}\label{eq:mle}
\theta^* = \arg\max_\theta \sum_{i} \sum_{j} \left(d^{'y_j}_{x_i} \ln p(y_j |x_i ; \theta) \right).
\end{equation}

 For the original hard target multi-classification problem, 
 the optimization equation from the cross-entropy loss is presented:

\begin{equation}\label{eq:ce.loss}
\theta^{*} =\arg\min_\theta \sum_{i} \sum_{j} d^{y_j}_{x_i} \log p(d^{y_j}_{x_i}|x_{i}; \theta)
\end{equation}


 The corresponding maximum likelihood function is shown as follows:
\begin{equation*}
\theta^* = \arg\max_{\theta} \sum_{i} \ln p(y(x_i) | x_i; \theta),
\end{equation*}
where $y(x_i)$ is the underlying true label for the $i$-t training instance.


For single-label problems, where the model only considers one label, it is necessary to compute the entropy of that particular label. 
Single-label learning is a special cases of the distribution of LS \citep{Geng2016}. LS can be seen
as a more general learning technique due to its relationship with traditional training methods.


Now we illustrate how we train the deep learning models with LS. The goal of model training is to iteratively optimize the parameter $\theta$ in order to find the optimal parameter $\theta^{*}$ that satisfies:
\begin{equation*}
\arg \min_{\theta} \sum_{i}KL(\boldsymbol{D}^{'}_{i}||P(\boldsymbol{y}|x_i; \theta)
\end{equation*}

The stochastic gradient descent (SGD) algorithm and its variants like Adam \citep{Kingma2015adam} are used to update the parameters. For each mini-batch, the updating formula is
\begin{equation}
\theta_{i+1} = \theta_{i} - \eta \cdot \delta_i
\end{equation}
where $\theta_{i+1}$ is the parameter vector at iteration $i+1$, $\theta_{i}$ is the parameter vector at iteration $i$, $\eta$ is the learning rate, which controls the step size of each update.
$\delta_i$ is the gradient of the loss function with respect to the parameter $\theta_{i}$.

\subsection{Selected Deep Learning Architectures}
TextCNN, BERT, and RoBERTa are three highly influential and successful deep learning architectures in natural language processing. 

{\bf TextCNN} \citep{kim2014convolutional}, as a pioneering architecture, introduced the idea of applying convolutional neural networks to text classification tasks. It leverages convolutional layers with various kernel sizes to capture local features in text data. TextCNN has demonstrated strong performance and efficiency in handling fixed-length input sequences \citep{Wang2022,Jiang2022}.

{\bf BERT} (Bidirectional Encoder Representations from Transformers) \citep{bert2018} brought about a breakthrough in language understanding. BERT introduced the concept of pre-training a transformer-based model on large amounts of unlabeled text data and fine-tuning it for downstream tasks. It showed remarkable success in a wide range of natural language processing tasks, including question answering, sentiment analysis, and named entity recognition.

{\bf RoBERTa} \citep{roberta2019}, building upon BERT's foundation, refined the pre-training process and achieved improved performance. It incorporated additional pre-training data and introduced modifications to the training objectives. RoBERTa demonstrated enhanced language representation capabilities and surpassed BERT's performance on various benchmarks and tasks.

\begin{figure*}[t]
    \centering
    \includegraphics[width=\textwidth]{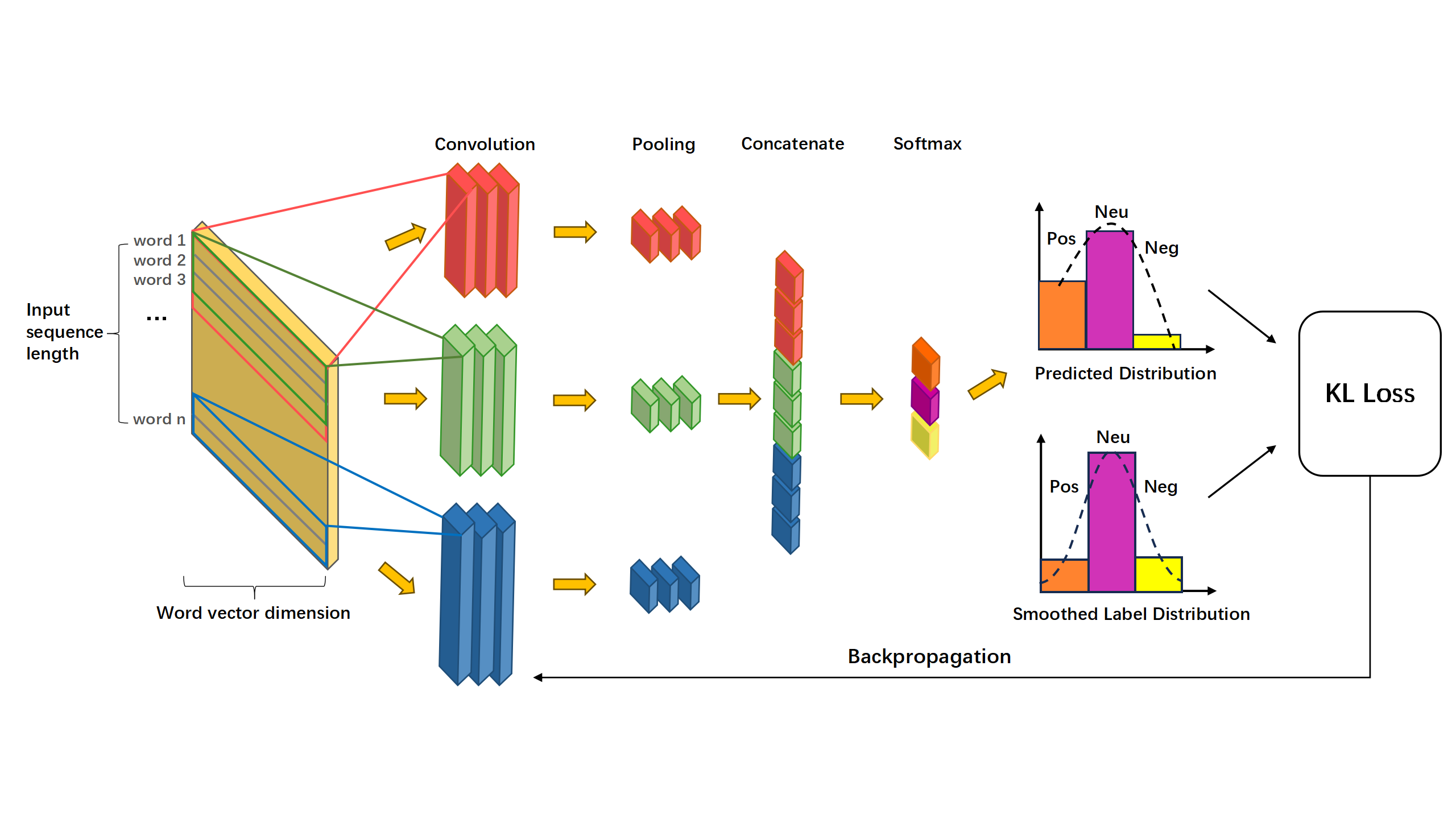}
    \caption{\label{fig:workflow}Workflow of our sentiment classification with label smoothing. Here the deep architecture is TextCNN, but the workflow for BERT and RoBERTa is similar.}
\end{figure*}

In our experiments, we investigate the effectiveness of LS across three aforementioned architectures.
The workflow of 
our method is shown in Fig. \ref{fig:workflow}, taking TextCNN as the example.

\section{Experimental Results and Analysis}\label{sec:exp}
We construct four LS predictive models incorporating varying degrees of label smoothing and assess their performance compared to the baseline models on eight datasets using TextCNN, BERT, and RoBERTa architectures. 
In additional to evaluating the effectiveness of LS methods, we also conduct in-depth analyses of how LS benefits text sentiment classification.

\subsection{Datasets}
We utilize publicly available datasets open-sourced in Kaggle and Huggingface. This experiment employs eight distinct sentiment analysis datasets, of which six are three-class and two are two-class. The diverse range of datasets enables us to evaluate the performance and generalizability of our models across various data types and scenarios.

\textbf{Twitter Financial News Sentiment (TFNS)} was collected using the Twitter API. It holds 11,932 documents annotated with 3 labels ('Positive', 'Neutral', 'Negatve') and comprises annotated finance-related tweets.

\textbf{Kaggle Financial Sentiment (KFS)} is a dataset of financial reviews annotated with 3 labels ('Positive', 'Neutral', 'Negatve'). These reviews cover topics such as stocks, investments, market analysis, corporate performance, technology, and real estate.

\textbf{Tweet Sentiment Extraction (TSE)} is a three-class dataset from a Kaggle contest. It has 27481 samples of a tweet and a sentiment label ('Positive', 'Neutral', or 'Negatve').

\textbf{Auditor Sentiment (AS)} gathers the auditor evaluations into one dataset. It contains thousands of sentences from English financial news, grouped into three categories by emotion ('Positive', 'Neutral', 'Negatve').

\textbf{Financial Phrasebank (FP)} consists of 4840 sentences from English language financial news categorised by sentiment. The dataset is divided into three classes ('Positive', 'Neutral', 'Negatve') by agreement rate of 5-8 annotators.

\textbf{Chatgpt Sentiment Analysis (CSA)} contains 10,000 pieces of data. The dataset gathers tweets about Chatgpt and Reflects people's perception of it. Emotions are divided into three categories ('Positive', 'Neutral', 'Negatve').

\textbf{Rotten Tomatoes Reviews (RTR)} consists of 10,662 processed sentences from Rotten Tomatoes movie reviews. It's a balanced two-class dataset with 5,331 positive and 5,331 negative.

\textbf{Sentiment140 (Sent140)} is a widely used sentiment analysis dataset created by researchers at Stanford University. It contains 1,600,000 tweets extracted using the twitter api. The tweets have been annotated to two classes ('Positive', 'Negatve'). 

\subsection{Model Configuration}
{\bf Models:}
Table \ref{tab:conf} presents the configuration of the models and loss functions used in the experiment. The LS models (LS1--LS4) employ different smoothing levels, with the smoothing parameter $\lambda$ increasing from 0.01 (LS1) to 0.1 (LS4). The original one-hot hard labels are used in the baseline methods, while for LS1 to LS4, the soft labels are utilized with different smoothing parameters. By gradually increasing the smoothing parameter from LS1 to LS4, we explore how label smoothing affects the model's accuracy and stability. All LS models utilize the KL divergence as the loss function. In contrast, the baseline models employ the cross entropy loss.

\begin{table*}[ht]
\centering
\caption{\label{tab:conf}Model smoothing levels and loss functions. The baseline methods take the original one-hot hard label, $\mathbf{y}$. We vary the smoothing parameter $\lambda$ to obtain four sets of soft labels $\mathbf{y}^{LS}$.}
\begin{tabular}{c l l l c}
\hline
   \hline
    Model &Smoothing para.(three/two-class)&Smoothed label (three-class)&Smoothed Label (two-class)& Loss\\
    \hline
    Baseline & $\lambda=0/0$ &$\mathbf{y}$ & $\mathbf{y}$ & Cross Entropy \\
    LS1 &  $\lambda=0.01/0.01$& $\mathbf{y}^{LS}=\mathbf{y}*0.97+0.01*\mathbf{1}$ & $\mathbf{y}^{LS}=\mathbf{y}*0.98+0.01*\mathbf{1}$ & KL Loss \\
    LS2 &  $\lambda=0.025/0.05$& $\mathbf{y}^{LS}=\mathbf{y}*0.925+0.025*\mathbf{1}$ & $\mathbf{y}^{LS}=\mathbf{y}*0.9+0.05*\mathbf{1}$ & KL Loss \\
    LS3 &  $\lambda=0.05/0.1$& $\mathbf{y}^{LS}=\mathbf{y}*0.85+0.05*\mathbf{1}$ & $\mathbf{y}^{LS}=\mathbf{y}*0.8+0.1*\mathbf{1}$ & KL Loss \\
    LS4 & $\lambda=0.1/0.15$& $\mathbf{y}^{LS}=\mathbf{y}*0.7+0.1*\mathbf{1}$ & $\mathbf{y}^{LS}=\mathbf{y}*0.7+0.15*\mathbf{1}$ & KL Loss \\
    \hline
    \hline
 \end{tabular}
\end{table*}

\begin{table*}[htb]
\centering
\caption{\label{tab:acc}The test accuracy of different classifiers on eight sentiment classification datasets with varying smoothing parameters.}
\tabcolsep=5pt
\begin{tabular}{c|c|c c c c c c c c}
\hline
\hline
Architecture & Algorithm & TFNS & KFS & TSE & AS & FP & CSA & RTR & Sent140\\
\hline
\hline
\multirow{5}{*}{BERT} 
& LS1 & \textbf{0.8769} & \textbf{0.7981} & \textbf{0.7917} & 0.8411 & 0.8930 & 0.7202 & \textbf{0.7820} & 0.7636 \\
& LS2 & 0.8744 & 0.7964 & 0.7912 & 0.8318 & 0.8955 & 0.7181 & 0.7780 & 0.7647 \\
& LS3 & 0.8710 & 0.7913 & 0.7881 & 0.8390 & \textbf{0.8968} & \textbf{0.7254} & 0.7780 & \textbf{0.7711} \\
& LS4 & 0.8740 & 0.7956 & 0.7889 & \textbf{0.8421} & 0.8955 & 0.7181 & 0.7860 & 0.7594 \\
\cline{3-10}
& Baseline & 0.8689 & 0.7896 & 0.7881 & 0.8256 & 0.8803 & 0.7149 & 0.7780 & 0.7626 \\
\hline
\multirow{5}{*}{TextCNN} & LS1 & \textbf{0.8237} & \textbf{0.6852} & \textbf{0.7074} & 0.7761 & 0.8382 & 0.7234 & 0.7860 & 0.7455 \\
& LS2 & 0.8233 & 0.6801 & 0.7029 & 0.7740 & 0.8357 & 0.7245 & \textbf{0.7920} & 0.7540 \\
& LS3 & 0.8216 & 0.6792 & 0.7066 & \textbf{0.7771} & 0.8369 & 0.7183 & 0.7780 & 0.7455 \\
& LS4 & 0.8204 & 0.6843 & 0.7029 & 0.7730 & 0.8344 & \textbf{0.7338} & 0.7860 & 0.7444 \\
\cline{3-10}
& Baseline & 0.8141 & 0.6851 & 0.7049 & 0.7647 & 0.8319 & 0.7090 & 0.7918 & \textbf{0.7626} \\
\hline
\multirow{5}{*}{RoBERTa} & LS1 & 0.8915 & \textbf{0.8375} & 0.7946 & \textbf{0.8689} & 0.9031 & 0.8101 & 0.8680 & 0.8588 \\
& LS2 & 0.8961 & 0.8263 & \textbf{0.8036} & 0.8617 & 0.9070 & 0.8029 & 0.8660 & 0.8524 \\
& LS3 & 0.8957 & 0.8246 & 0.7999 & 0.8658 & \textbf{0.9108} & \textbf{0.8163} & 0.8720 & 0.8610 \\
& LS4 & \textbf{0.8978} & 0.8298 & 0.8024 & 0.8638 & 0.9045 & 0.8122 & \textbf{0.8780} & 0.8706 \\
\cline{3-10}
& Baseline & 0.8769 & 0.8315 & 0.7991 & 0.8566 & 0.8994 & 0.8039 & 0.8720 & \textbf{0.8759} \\
\hline
\hline
\end{tabular}
\end{table*}

\begin{figure*}[htb]
\centering
\caption{\label{fig:valid.acc}Accuracy on the validation set at different epochs using BERT with varying smoothing parameters.}
\includegraphics[width=\textwidth]{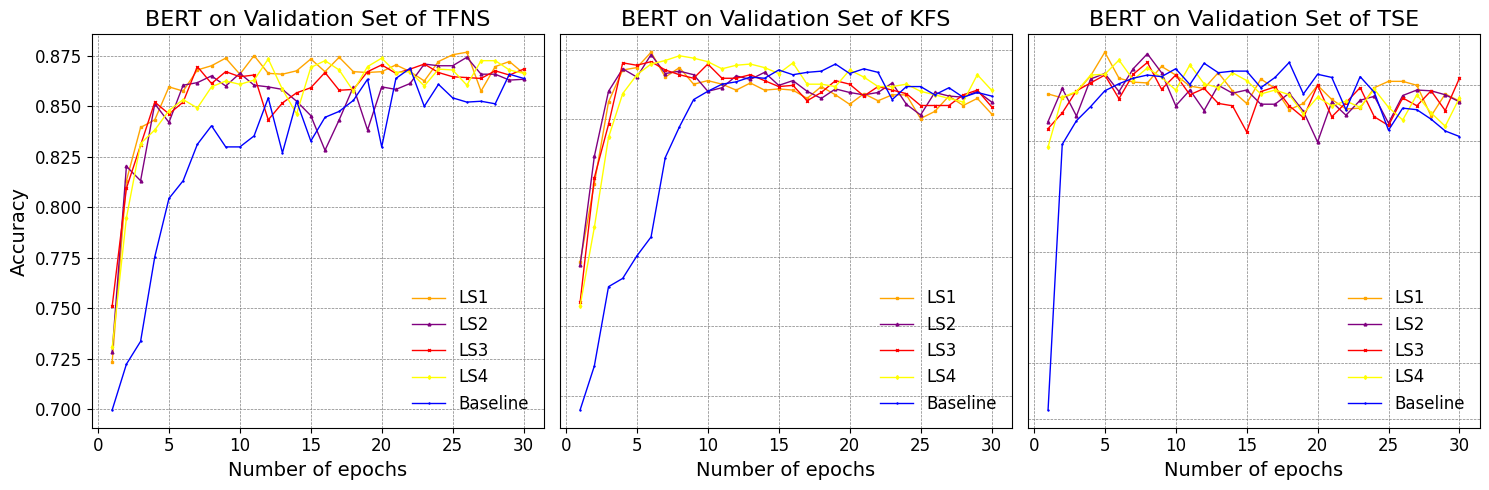}
\end{figure*}



\subsection{Results}
{\bf Performance:} We employ accuracy as the main metrics to evaluate the performance of methods on eight datasets. 
The results produced by BERT, TextCNN and RoBERTa architectures are presented in Table \ref{tab:acc}. 
TextCNN models are trained from scratch with random initializations, while BERT and RoBERTa models are finetuned on sentiment analysis datasets with pre-trained weights.
On the BERT architecture, LS models consistently outperform the baseline model across multiple datasets. Specifically, LS1 achieves the highest accuracy in most cases, closely followed by LS3 and LS4. This indicates that incorporating LS with varying degrees of LS can significantly improve classification accuracy, particularly for binary and ternary classification datasets.

Moving to the TextCNN architecture, LS models exhibit superior performance on three-category classification datasets compared to binary classification datasets. LS1 and LS3 consistently outperform the other models in terms of accuracy. This highlights the effectiveness of LS when implemented in the TextCNN architecture for enhancing classification outcomes. 

On the RoBERTa architecture, LS models demonstrate competitive performance, LS1 and LS4 achieving the highest accuracy in most cases. This suggests that LS can effectively leverage the strengths of the RoBERTa architecture to achieve improved classification accuracy.

The experimental results demonstrate that LS models, when integrated into different architectures, consistently outperform the baseline model. LS1, LS3, and LS4 consistently demonstrate superior accuracy, indicating the efficacy of LS in enhancing classification performance.

\begin{figure*}
\centering
\begin{subfigure}{0.5\textwidth}
    \includegraphics[width=\textwidth]{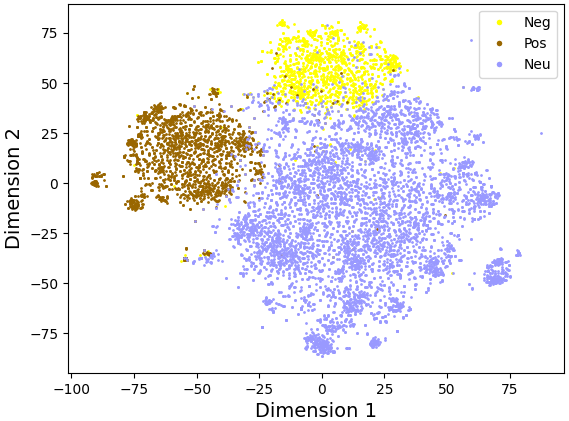}
    \caption{t-SNE plot from BERT baseline}
    \label{fig:baseline}
\end{subfigure}
\hfill
\begin{subfigure}{0.49\textwidth}
    \includegraphics[width=\textwidth]{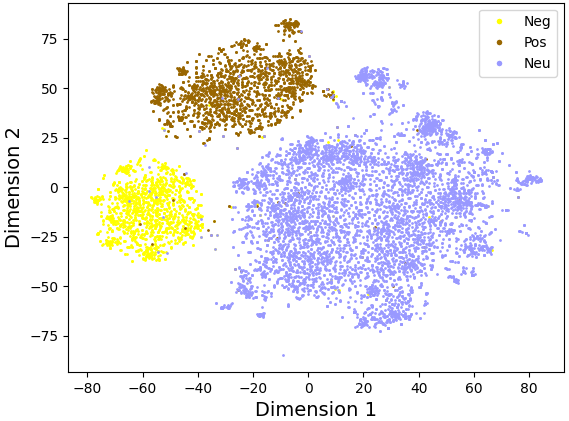}
    \caption{t-SNE plot from BERT with LS1}
    \label{fig:ls2}
\end{subfigure}
\hfill
\begin{subfigure}{0.5\textwidth}
    \includegraphics[width=\textwidth]{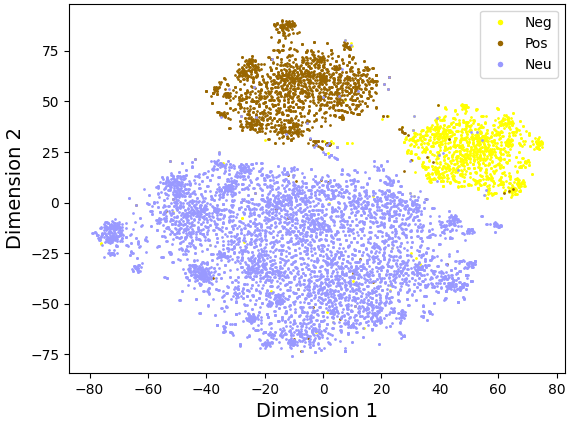}
    \caption{t-SNE plot from BERT with LS2}
    \label{fig:ls4}
\end{subfigure}
\begin{subfigure}{0.49\textwidth}
    \includegraphics[width=\textwidth]{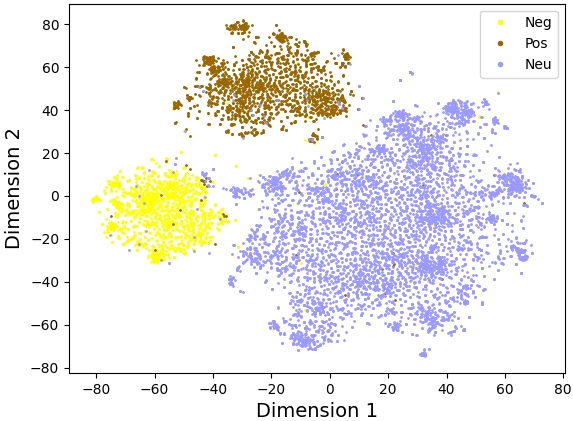}
    \caption{t-SNE plot from BERT with LS3}
    \label{fig:ls5}
\end{subfigure}
\caption{t-SNE plots from BERT baseline versus label smoothing methods.}
\label{fig:tsne}
\end{figure*}

\subsection{Analysis}
In this section, we attempt to analyze why and how LS benefits the text sentiment classification.

{\bf Higher Accuracy}: As shown in Table \ref{tab:acc}, LS methods achieve higher accuracy than baseline models. The impact of label distribution can be profound, as it effectively addresses certain challenges in sentiment classification. Label distribution helps alleviate overconfidence in model predictions and reduces sensitivity to noisy labels, resulting in improved robustness and generalization performance.

Furthermore, in sentiment classification, emotions are not always purely positive or negative. There can be instances where a dominant emotion encompasses other subtle emotions. LS models excel at capturing and representing such phenomena, allowing for a more nuanced understanding and interpretation of sentiment. By considering a broader spectrum of emotions and incorporating LS techniques, LS models offer a more comprehensive and accurate representation of sentiment classification, leading to improved performance in sentiment analysis tasks.

{\bf Acceleration of Convergence}: We further analyzed the convergence performance of deep learning architectures during the training of sentiment classifiers. The accuracy curves of BERT are presented in Figure \ref{fig:valid.acc}, depicting the performance of the models on the validation sets at each training epoch.
 It can be observed that LS models implemented on BERT exhibit significantly faster convergence compared to the baseline model, and can even achieve relatively high accuracy within the initial few epochs of training. This also indicates that LS is capable of effectively saving computational resources, reducing the training time, and improving overall efficiency.


{\bf Reduce Overfitting}: LS and KL divergence can help reduce overfitting compared to cross-entropy for several reasons. Cross-entropy loss assigns high confidence to a single target class and penalizes all other classes. This can lead to overconfident predictions, especially when the training data is limited or imbalanced. In contrast, LS considers the entire label distribution, allowing the model to capture the relationships between classes and reduce the risk of overfitting to individual examples. After applying the LS method, during the training phase, the loss does not decrease too rapidly when the predictions are correct, and it does not penalize too heavily when the predictions are wrong. This prevents the network from easily getting stuck in local optima and helps to mitigate overfitting to some extent. Additionally, in scenarios where the classification categories are closely related, the network's predictions are not excessively absolute.

{\bf Representation Learning}: 
LS can produce better hidden representations for texts than baseline methods. The effects of LS can be reflected in the hidden layer of the Model. Figure \ref{fig:tsne} visualizes similarities and differences between data points by mapping high-dimensional features into two-dimensional spaces and depicts the last layer of BERT models with different smoothing levels. The data points from baseline BERT model overlap as shown in Fig. \ref{fig:baseline}, while datapoints from LS methods are linearly separable as shown in Fig \ref{fig:ls2}--\ref{fig:ls5}. 

\section{Conclusion}
In this paper, we delve into the application of LS in text sentiment classification, a key problem of natural language processing. Our experiments demonstrate that by tuning the smoothing parameters, LS can achieve improved prediction accuracy across multiple deep learning architectures and various test datasets. We conduct fine-grained analysis to check how LS benefits the task. Our main findings are: 1.) LS can boost the convergence of deep models in both training from scratch and fine-tuning modes; 2.) LS can produce hidden representations for texts that
make the data of different labels more easily separable.
However, given the inherent emotional bias in different texts, constructing more precise textual sentiment distribution labels to capture subtle variations in human emotions remains a challenging task. This highlights the need for further research and development in this area to fully harness the potential of LS in enhancing the performance and robustness of sentiment classification models.

\bibliographystyle{IEEEtranN}
\bibliography{IEEEabrv,myref}

\end{document}